\documentclass[runningheads]{llncs}
\pdfoutput=1
\usepackage{graphicx}
\usepackage{bbold}
\usepackage[dvipsnames]{xcolor}
\usepackage{hyperref}
\hypersetup{
	colorlinks,
	linkcolor={red!50!black},
	citecolor={blue!50!black},
	urlcolor={blue!80!black}
}
\usepackage{booktabs}
\usepackage{multirow}
\usepackage{adjustbox}
\usepackage{capt-of}
\usepackage{cite}

\newcommand{\ie}{\textit{i}.\textit{e}.}
\newcommand{\eg}{\textit{e}.\textit{g}.}
\newcommand{\etc}{\textit{etc}.}

\usepackage[width=122mm,left=12mm,paperwidth=146mm,height=193mm,top=12mm,paperheight=217mm]{geometry}

\begin{document}
\title{Self-supervised Contrastive Video-Speech Representation Learning for Ultrasound}
\titlerunning{Self-supervised Contrastive Video-Speech Representation Learning}
%
\author{Jianbo Jiao\inst{1}
\and
Yifan Cai\inst{1} \and
Mohammad Alsharid\inst{1} \and
Lior Drukker\inst{2} \and
Aris T. Papageorghiou\inst{2} \and
J. Alison Noble\inst{1}
}
%
\authorrunning{J. Jiao et al.}
%
\institute{Department of Engineering Science, University of Oxford, Oxford, UK\\
\email{jianbo.jiao@eng.ox.ac.uk}
\and
Nuffield Department of Women’s \& Reproductive Health, University of Oxford, UK}
\maketitle              
\begin{abstract}
    In medical imaging, manual annotations can be expensive to acquire and sometimes infeasible to access, making conventional deep learning-based models difficult to scale. As a result, it would be beneficial if useful representations could be derived from raw data without the need for manual annotations. In this paper, we propose to address the problem of self-supervised representation learning with multi-modal ultrasound video-speech raw data. For this case, we assume that there is a high correlation between the ultrasound video and the corresponding narrative speech audio of the sonographer. In order to learn meaningful representations, the model needs to identify such correlation and at the same time understand the underlying anatomical features. We designed a framework to model the correspondence between video and audio without any kind of human annotations. Within this framework, we introduce cross-modal contrastive learning and an affinity-aware self-paced learning scheme to enhance correlation modelling. Experimental evaluations on multi-modal fetal ultrasound video and audio show that the proposed approach is able to learn strong representations and transfers well to downstream tasks of standard plane detection and eye-gaze prediction.
\end{abstract}
\section{Introduction}
Deep learning-based medical image analysis approaches rely heavily on annotated training data, which limits the progress in medical image analysis if a large dataset has to be manually annotated for every new task.
Extracting meaningful representations directly from unlabelled data is therefore an important and interesting sub-topic in learning-based medical image analysis.

Several approaches in the literature have been explored to deal with the problem of learning from unlabelled data, which is usually termed as ``self-supervised representation learning'' (or unsupervised learning in some works).
The common practice is to pre-train a model on unlabelled data according to a pretext task and then to fine tune the model with some specific target tasks to evaluate the learned representations.
Typical pretext tasks include colourisation~\cite{zhang2016colorful}, and rotation prediction~\cite{gidaris2018unsupervised} for images; and tracking~\cite{wang2015unsupervised}, temporal ordering~\cite{lee2017unsupervised} for videos.
Some recent studies propose to learn representations by contrastive predictive coding~\cite{oord2018representation,henaff2019data} or contrastive learning~\cite{he2020momentum,chen2020simple}, and showed a powerful learning capability.
There are also works have considered medical images, \eg, predicting the distance between patches~\cite{spitzer2018improving}, Rubik's cube recovery~\cite{zhuang2019self} {and anatomy-aware joint reasoning~\cite{jiao2020self}}.
However, the above-mentioned approaches are single modality.
Some recent approaches have investigated learning from natural audio and video modalities~\cite{arandjelovic2017look,Owens_2018_ECCV}, where the pretext task is designed as video-audio alignment.
In this case the audio and video are assumed to be in dense correspondence. Such multi-modal learning has not been explored for medical data before.
Audio data rarely exists for medical images and even when available, it is mostly narrative diagnosis/interpretation speech, which has a sparse correlation with the visual data, making the task rather challenging.

In this paper, we propose to address the problem of self-supervised cross-modal representation learning for ultrasound video with corresponding narrative speech audio, both of which are captured on-the-fly without any manual annotations.
Unlike other medical imaging modalities that are with clear anatomical structures (\eg, MRI and CT), ultrasound video is much more difficult to interpret by eye for humans although experts are adept at interpreting anatomy in the acoustic patterns.
As a result, learning representations automatically from unlabelled ultrasound data is rather challenging.
On the other hand, in our case we have synchronised narrative speech from the sonographer accompanied with the ultrasound video.
The basic assumption here is that by leveraging cross-modal correlations, a useful representation can be learned.
To this end, we propose to learn the anatomical representations from ultrasound video-speech data by identifying the affinity (\ie, association strength) between the two modalities.
Specifically, we first randomly sample video-speech samples from our dataset, from which positive and negative pairs are generated.
Unlike prior works that use straightforward simple training pairs, in this work we instead leverage hard-negative as well as hard-positive pairs for training, in order to force the model to \emph{learn harder}.
Additionally, we further introduce cross-modal contrastive learning to encourage a positive pair to have strong affinity while a negative pair will have weaker affinity in a projected shared latent space.
In our diagnostic narrative speech audio, we observe that background noise (\eg, button clicks, air conditioning) and unrelated conversation (\eg, about the weather, travel \etc) exist in the raw data, which degrades the representation learning.
To mitigate this problem, we propose an affinity-aware self-paced learning curriculum over the representation learning process.

To evaluate the proposed self-supervised learning framework, we consider two ultrasound-related downstream tasks: standard plane detection and eye-gaze saliency prediction.
Experimental results presented show that the proposed approach significantly improves the performance of downstream tasks without referring to any manual annotations.
The experiments also reveal that by leveraging speech data, useful representations can be learned when transferring to other tasks, outperforming single-modal learning methods.

The main contributions of this paper are summarised as follows:
\begin{itemize}
   \item We propose, to our knowledge, the first self-supervised video-speech representation learning approach for ultrasound data.
   \item We introduce cross-modal contrastive learning and affinity-aware self-paced learning for ultrasound video-speech cross-modal representation learning.
   \item The proposed approach is demonstrated to be effective for two downstream tasks.
\end{itemize}

\section{Method}
\begin{figure}[t]
    \centering
   \includegraphics[width=\textwidth]{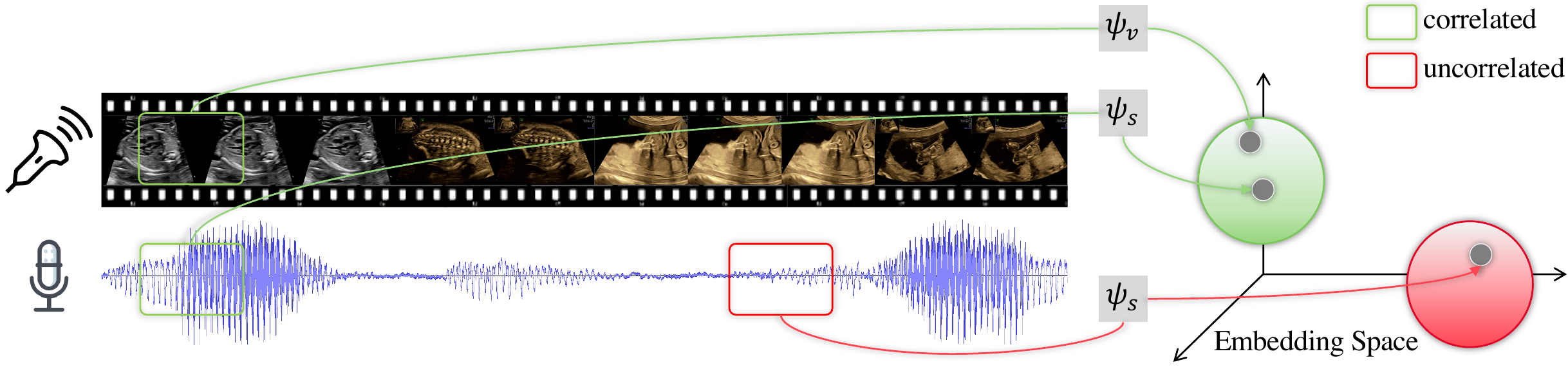}
   \caption{Main idea of the proposed self-supervised video-speech representation learning framework. A model is trained to identify whether a sampled video-speech pair is anatomically correlated, and at the same time encourage the projected embeddings from correlated pair to lie on the same anatomical sphere (\eg, the green one).}
    \label{fig:overview}
\end{figure}

In this section, we present the proposed self-supervised contrastive video-speech representation learning approach in detail.
The main idea of the proposed approach is to build the correlation between video and narrative speech data by both explicit optimisation and implicit regularisation.
In this paper, we demonstrate the idea with fetal ultrasound video synchronised with corresponding speech.
A novel self-supervised learning framework is proposed accordingly.
Fig.~\ref{fig:overview} illustrates the main idea of the proposed framework.

\subsection{Video-Speech Representation Learning}
Speech audio, as an additional modality to visual data (\eg, images and videos), is recorded on-the-fly without external annotations. The idea is to use the discriminative power of audio to understand visual data where ambiguity may exist.
The basic assumption to learn a cross-modal (ultrasound video and corresponding speech from sonographer in our case) representation is that the two modalities share similar anatomical meanings at the same timestamp.
In our case, if a model can successfully identify whether a video clip and a speech clip are correlated or not, it has learned to understand the underlying anatomical representations.

\begin{figure}[t]
    \centering
    \includegraphics[width=\textwidth]{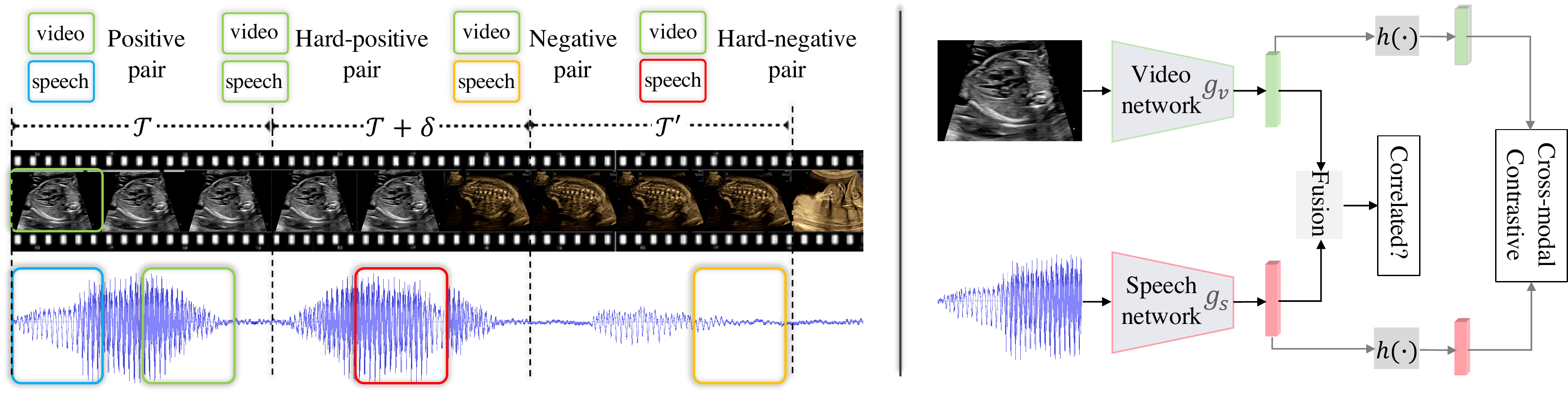}
    \caption{Left: proposed training pair sampling scheme, where the hard-positive and hard-negative are illustrated. Right: proposed framework for self-supervised video-speech representation learning.}
    \label{fig:framework}
\end{figure}

Based on the above assumption, we design a self-supervised learning framework to extract cross-modal representations with ultrasound video-audio data.
Specifically, we randomly sample positive and negative pairs from the original video-speech data, where a positive pair indicates that the considered video and speech are correlated, while a negative pair is uncorrelated.
A deep model is then trained to identify such positive/negative pairs accordingly, resembling a classification problem.
Unlike natural image video and its corresponding audio where highly-dense correlations present (\eg, playing the violin, cooking), the speech audio from a sonographer and the medical ultrasound video are sparsely correlated and narratively presented, making the correlation identification more challenging.
To address this issue, we first propose to force the model to \emph{learn harder} by sampling hard-negative and hard-positive pairs (as illustrated in Fig.~\ref{fig:framework} left), so as to learn a more strongly correlated representations.
Suppose the ultrasound video and speech clips at time interval $\mathcal{T}$ are $\mathcal{V_T}$ and $\mathcal{S_T}$, the speech clip at a shifted time interval $\mathcal{T'}$ is $\mathcal{S_{T'}}$, $(\mathcal{V_T}, \mathcal{S_T})$ is considered as a positive pair and $(\mathcal{V_T}, \mathcal{S_{T'}})$ a negative pair.
Instead of sampling $\mathcal{T'}$ from a different scan sequence or a diverse anatomy sub-sequence, we force the learning process to be constrained by sampling $\mathcal{T'}$ from a nearby segment. This does not have to be for a different anatomy as $\mathcal{T}$, \ie, $\mathcal{T'}=\mathcal{T}+\delta$ where $\delta<\mathcal{D}$ is a short random timestep and $\mathcal{D}$ is the shift range.
Furthermore, we sample ultrasound video frames and speech clips to increase the generalisability: $\{v_t, s_t, s_{t'}|v\in\mathcal{V},s\in\mathcal{S},t\in\mathcal{T},t'\in\mathcal{T'}\}$.
In terms of the positive pair, we also make it \emph{harder} by perturbing the alignment within $\mathcal{T}$ so that $v_t$ and $s_t$ do not have to be exactly at the same timestamp $t$.
Additionally, we randomly sample a group of positive/negative pairs within each interval to represent the whole clip.
Finally, the hard-positive and hard-negative pairs used for training are defined as $\{(v_{ik}, s_{jk}) | k\in[1,K]\}$ and $\{(v_{ik}, s_{t'k}) | k\in[1,K]\}$ respecctively, where $i,j\in\mathcal{T}$ and $K$ is the number of sampling groups.
An illustration of the proposed pair sampling scheme when $K=1$ is shown in Fig.~\ref{fig:framework} (left).

With the above-mentioned sampling scheme, training pairs are fed into deep networks to extract the corresponding features.
Suppose the ultrasound video and speech subnetworks as $g_v$ and $g_s$, the extracted features are $g_v(v_t)$ and $g_s(\eta(s_t))$ where $\eta(\cdot)$ is a speech pre-processing function that converts the 1D signal to a 2D spectrogram (more details see Sec.~\ref{sec:expset}).
The video and speech features are fused for the correlation decision with a fusion function $f(v_t,s_t)=g_v(v_t)\oplus g_s(\eta(s_t))$ where $\oplus$ represents feature concatenation.
The correlation decision task is then modelled as a classification problem with the objective:
\begin{equation}
    \mathcal{L}_{cls}=-\frac{1}{N}\sum\nolimits_{n=1}^N\sum\nolimits_i^C c_i^n log(f(v_t,s_t)_i^n),
\end{equation}
where $N$ is the total number of samples while $C=2$ for our binary classification problem, $c$ is the label indicating if the inputs are correlated or not.

\subsection{Cross-modal Contrastive Learning}
We observe that in the speech data, there exist medical-unrelated contents (\eg, random talk), which deteriorate correlation learning.
To address this, we introduce cross-modal contrastive learning, in addition to the afore-mentioned classification objective.
The key idea here is to encourage the representations of video and speech from a (hard-)positive pair to be similar while repelling the (hard-)negative pair, at a projected embedding space.
Specifically, before the feature fusion, $g_v(v_t)$ and $g_s(\eta(s_t))$ are further projected to a latent space by function $h(\cdot)$, where the cross-modal contrastive objective is applied.
Suppose $y_v=h(g_v(v_t)), y_s=h(g_s(\eta(s_t)))$ are the projected embeddings, the cross-modal contrastive objective is defined as:
\begin{equation}
    \mathcal{L}_{cont}=-log\frac{e^{sim(y_v,y_s)} - e^{sim(y_v,y_{s'})}}{\sum\nolimits_{k=1}^{N}\mathbb{1}_{[k\neq v]}e^{sim(y_v,y_k)}},
\end{equation}
where $\mathbb{1}\in\{0,1\}$ and $y_{s'}=h(g_s(\eta(s_{t'})))$. Function $sim(a,b)=a^\top b$ measures the similarity between two vectors.
An illustration of the additional cross-modal contrastive learning is shown in Fig.~\ref{fig:framework} (right).

\subsection{Affinity-Aware Self-paced Learning}

The microphone used to record the speech is placed in an open environment without any specific voice filtering, which means that all the sound is recorded including background noise (\eg, air conditioner, button-click), in addition to the main narrative content.
Since our main focus is the speech with meaningful descriptions, the noise inevitably affects representation learning.
To this end, we further propose a self-paced learning curriculum based on the affinity between video and speech data.
Specifically, we divide the multi-modal data into different affinity levels and perform specific learning schemes accordingly.
For simplicity, here we choose two affinity levels, \ie, low-affinity and high-affinity, where low-affinity refers to the speech audio mostly consisting of noise and the rest is called high-affinity.
The affinity is automatically detected according to an energy-based voice activity detection algorithm~\cite{sakhnov2009approach}.

The proposed representation learning approach is only performed on the high-affinity fragments.
In terms of the low-affinity data, instead of directly discarding it, we propose to leverage the video data which provides visual clues for representation learning.
As a result, we include an additional pretext task to extract representations from the whole video data.
Inspired by~\cite{lee2017unsupervised}, we randomly shuffle the frames in a video clip and train the model to predict the correct order.
The assumption here is that the model can correct the shuffled frame order only if it understands the anatomical information within the video clip.
Four frames are used to construct a clip for this task and the forward and backward sequences are considered to be the same (\eg, 0-1-2-3 \emph{v.s.} 3-2-1-0).
Therefore, this pretext task is modelled as a 12-category classification problem with cross-entropy objective $\mathcal{L}_{ord}$.
To avoid model cheating, the fan-shape of the ultrasound images is cropped out, keeping only the inner part.

\subsection{Implementation and Training}
The proposed model framework is illustrated in Fig.~\ref{fig:framework} (right).
In terms of the backbone network architecture, we choose the ResNeXt-50~\cite{xie2017aggregated} with Squeeze-and-Excitation module~\cite{hu2018squeeze} and dilated convolutions~\cite{yu2015multi}.
The video and speech subnetworks share the same architecture but are optimised separately with a joint objective function as defined in Eq.~\ref{eq:joint}:
\begin{equation}
    \label{eq:joint}
    \mathcal{L}=\alpha\mathcal{L}_{cls}+\beta\mathcal{L}_{cont}+\gamma\mathcal{L}_{ord},
\end{equation}
where $\alpha, \beta, \gamma$ are weighting parameters and are empirically {determined} to be equal.
The projection function $h(\cdot)$ is achieved by a multilayer perceptron (MLP) with a hidden layer and non-linear activation.
The models are trained with the SGD optimizer with momentum set to 0.9 and weight decay as $5\times 10^{-4}$.
The learning rate is initialised to $10^{-3}$ and divided by 10 for every 20 epochs.
The whole model is trained for 80 epochs. Gradient clipping is applied and the batch size is set to 32.
The sampling group number $K=2$ due to memory limitation and the interval skip range $\mathcal{D}=5$.
The whole framework is implemented using PyTorch on an NVIDIA RTX 2080Ti.

\section{Experiments and Analysis}

\subsection{Data and Experimental Settings}\label{sec:expset}
The data used in this work is from a routine clinical fetal ultrasound dataset~\footnote[1]{{UK Research Ethics Committee Reference 18/WS/0051}.} with the scanned video and corresponding real-time speech as well as eye-gaze data from sonographers.
In total, we have 81 scans with speech data.
On average, each video scan is about 55,000 frames with frame rate of 30 fps.
Each video clip $\mathcal{T}$ consists of 60 consecutive frames {and we have 73,681 clips for model training}.
When sampling a training pair within each video clip, we extract 0.6s of the corresponding speech data and resample it to 24kHz.
The speech is then converted to a 2D log-spectrogram representation of size $256\times256$, using a short-time Fourier transform (STFT) with 256 frequency bands, 10ms window length and 5ms hop length.
Two downstream tasks are included for the learned representation evaluation, where we use 135 scans with three-fold cross-validation (90/45 for train/test) and each scan is temporally down-sampled at a rate of 8.

\begin{table}[t]
    \begin{minipage}{0.49\textwidth}
  \caption{Evaluation results on standard plane detection (mean$\pm$std.[\%]). Best performance is marked in \textbf{bold}. Note the \textcolor{gray}{methods} on the right side are fully-supervised using external annotations.}
  \adjustbox{max width=\textwidth}{
  \centering
  \begin{tabular}{@{}l|ccc|ccc@{}}
    \toprule
    & Rand.Init. & Video & Ours & ImageNet Init. & SonoNet \\
    \midrule
    Precision & 70.4{\small$\pm$2.3} & 71.9{\small$\pm$2.0} & \textbf{72.7}{\small$\pm$1.8} & \textcolor{gray}{74.6{\small$\pm$1.8}} & \textcolor{gray}{82.3{\small$\pm$1.3}} \\
    Recall & 64.9{\small$\pm$1.6} & 71.7{\small$\pm$3.5} & \textbf{73.3}{\small$\pm$2.4} & \textcolor{gray}{71.2{\small$\pm$1.9}} & \textcolor{gray}{87.3{\small$\pm$1.1}} \\
    F1-score & 67.0{\small$\pm$1.3} & 71.5{\small$\pm$2.4} & \textbf{72.6}{\small$\pm$1.7} & \textcolor{gray}{72.5{\small$\pm$1.8}} & \textcolor{gray}{84.5{\small$\pm$0.9}} \\
    \bottomrule
  \end{tabular}
  }
  \label{tab:spd}
\end{minipage}
\hfill
\begin{minipage}{0.49\textwidth}
    \centering
    \includegraphics[width=\textwidth]{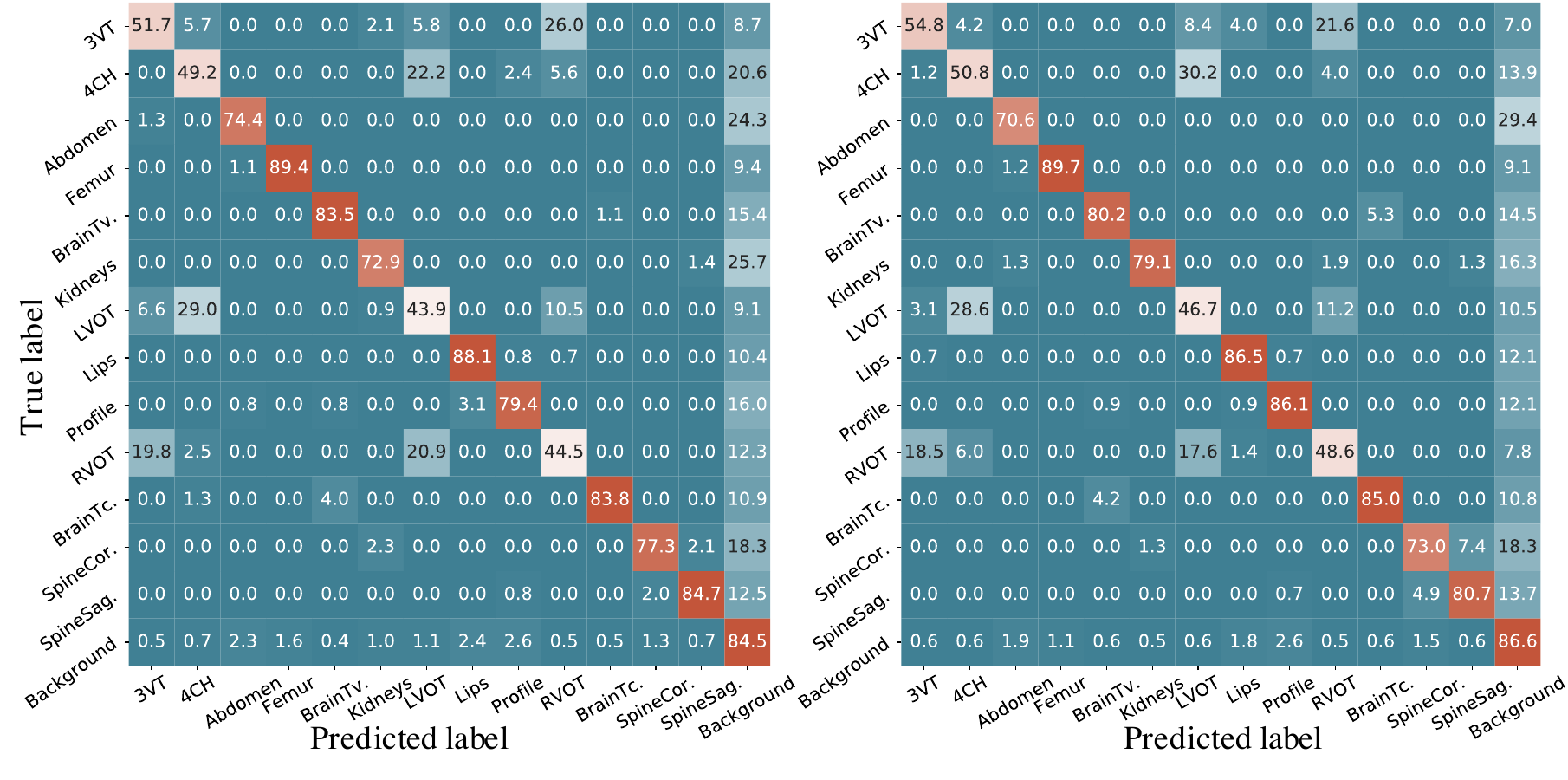}
    \captionof{figure}{Confusion matrix on standard plane detection. Left: \emph{Video}. Right: \emph{Ours}. (Best viewed in digital form.)}
    \label{fig:CM}
\end{minipage}
\end{table}

\subsection{Standard Plane Detection}\label{sec:spd}
To evaluate the learned representations, we first perform transfer learning on the standard plane detection task by fine-tuning the pre-trained weights from pretext tasks.
We use 14 categories of heart three-vessel and trachea view (3VT), heart four-chamber view (4CH), abdomen, femur, brain transventricular plane (BrainTv.), kidneys, heart left ventricular outflow tract (LVOT), lips, profile, heart right ventricular outflow tract (RVOT), brain transcerebellum plane (Brai-nTc.), spine coronal plane (SpineCor.), spine sagittal plane (SpineSag.) and background.
Standard plane labels are obtained from the dataset mentioned above.
Initialisation from random weights, pre-trained weights only on video data (by the aforementioned frame order prediction task), pre-trained weights on ImageNet~\cite{russakovsky2015imagenet} and weights pre-trained with SonoNet~\cite{baumgartner2017sononet} are included for comparison.
Quantitative results are presented in Table~\ref{tab:spd}.
We see that the proposed self-supervised cross-modal representation approach performs better than the other alternative solutions on average.
It also reveals that by leveraging the speech data with consideration of its correlation to the visual data, better performance is achieved, indicating stronger representations are learned.
Note that our learned representation performs better than the ImageNet initialisation for most metrics, which also suggests that representation extracted from natural images do not transfer very well to medical data.
In addition, we show the label confusion matrix in Fig.~\ref{fig:CM}.
It can be observed that our approach performs well on each category except the fine-grained ones like 3VT, 4CH, LVOT and RVOT (different views of fetal heart), which are challenging for sonographers too.
When compared to the video-only representations (Fig.~\ref{fig:CM} left), we can see that our approach improves on almost all the categories, especially the abovementioned fine-grained ones.
This is mainly due to the incorporation of visual-speech correlation where the additional speech representation reduces the ambiguity that exists in video data.

\subsection{Eye-Gaze Saliency Prediction}
Since our dataset contains simultaneous eye-gaze tracking data from sonographers, in addition to the standard plane detection task as in Sec.~\ref{sec:spd}, we further evaluate the effectiveness of the learned representations on a regression-based task, namely eye-gaze saliency prediction.
Similarly, we load the pre-trained weights and fine-tune on the downstream task.
A similar network architecture is used, with only the last layers modified to predict a 2D saliency map.
Following~\cite{bylinskii2018different}, we use the KL divergence (KL), normalised scanpath saliency (NSS), area under curve (AUC), correlation coefficient (CC) and similarity (SIM) as the evaluation metrics.
Quantitative and qualitative results are shown in Table~\ref{tab:sal} and Fig.~\ref{fig:salvis} respectively, from which we can see that our approach again outperforms the alternative solutions, and even better than the approaches (ImageNet Init. and SonoNet) that were pre-trained with manual annotations.

\begin{table}[t]
    \begin{minipage}{0.616\textwidth}
  \caption{Quantitative evaluation on eye-gaze saliency prediction. Best performance is marked in \textbf{bold}.}
  \centering
  \adjustbox{max width=\textwidth}{
  \begin{tabular}{@{}l|ccccc@{}}
    \toprule
    & KL$\downarrow$ & NSS$\uparrow$ & AUC$\uparrow$ & CC$\uparrow$ & SIM$\uparrow$\\
    \midrule
    Rand.Init. & 3.94{\small$\pm$0.18} & 1.47{\small$\pm$0.24} & 0.90{\small$\pm$0.01} & 0.12{\small$\pm$0.02} & 0.05{\small$\pm$0.01}\\
    Video & 3.57{\small$\pm$0.10} & 1.86{\small$\pm$0.12} & 0.91{\small$\pm$0.01} & 0.15{\small$\pm$0.01} & 0.08{\small$\pm$0.01} \\
    Ours & \textbf{3.05}{\small$\pm$0.04} & \textbf{2.68}{\small$\pm$0.05} & \textbf{0.95}{\small$\pm$0.00} & \textbf{0.22}{\small$\pm$0.00} & \textbf{0.11}{\small$\pm$0.00} \\
    \midrule
    ImageNet Init. & \textcolor{gray}{3.95{\small$\pm$0.28}} & \textcolor{gray}{1.72{\small$\pm$0.25}} & \textcolor{gray}{0.89{\small$\pm$}0.02} & \textcolor{gray}{0.14{\small$\pm$}0.02} & \textcolor{gray}{0.08{\small$\pm$}0.01} \\
    SonoNet & \textcolor{gray}{3.14{\small$\pm$0.02}} & \textcolor{gray}{2.62{\small$\pm$0.03}} & \textcolor{gray}{0.94{\small$\pm$0.00}} & \textcolor{gray}{0.21{\small$\pm$0.00}} & \textcolor{gray}{0.12{\small$\pm$0.00}} \\
    \bottomrule
  \end{tabular}
  }
  \label{tab:sal}
\end{minipage}
\hfill
\begin{minipage}{0.39\textwidth}
    \centering
    \includegraphics[width=\textwidth]{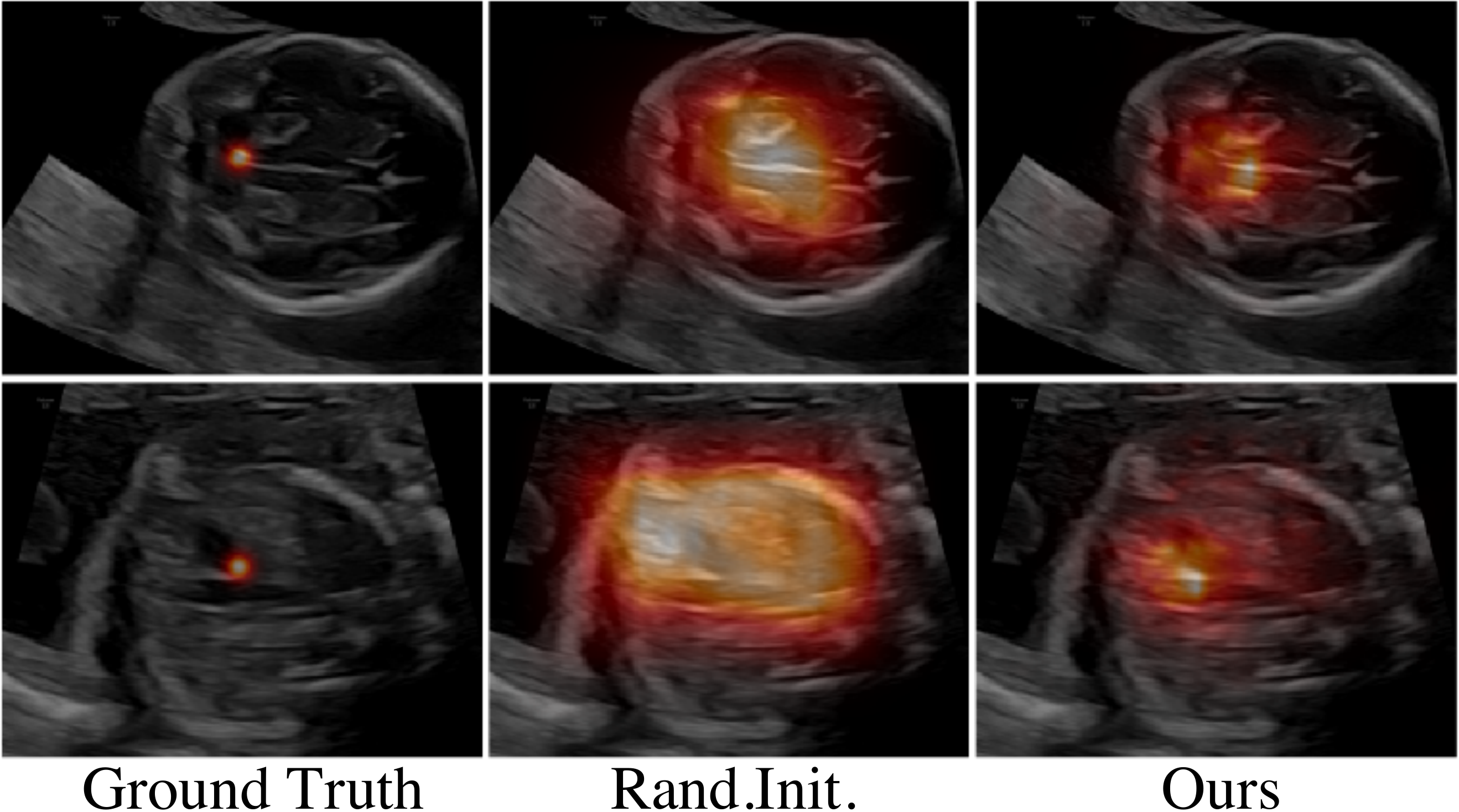}
    \captionof{figure}{Qualitative performance on eye-gaze saliency prediction.}
    \label{fig:salvis}
\end{minipage}
\end{table}

\subsection{Ablation Study}
To better understand the effectiveness of the proposed learning strategies, we present an ablation study with corresponding performance shown in Table~\ref{tab:abl}.
We take the video-speech learning approach without cross-model contrastive learning (\emph{CM.Contra.}) and affinity-aware self-paced learning as baseline.
We can see from Table~\ref{tab:abl} that when including the cross-modal contrastive learning, the performance is improved by a large margin.
Adding the affinity-aware learning scheme (\emph{Ours}), further improves the model.

\begin{table}
  \caption{Ablation study on each of the proposed strategies for two downstream tasks.}
  \centering
  \adjustbox{max width=\textwidth}{
  \begin{tabular}{@{}l|ccc|ccccc@{}}
    \toprule
    & Prec.[\%]$\uparrow$ & Rec.[\%]$\uparrow$ & F1[\%]$\uparrow$ & KL$\downarrow$ & NSS$\uparrow$ & AUC$\uparrow$ & CC$\uparrow$ & SIM$\uparrow$\\
    \midrule
    Baseline & 70.7{\small$\pm$3.6} & 70.2{\small$\pm$2.4} & 69.8{\small$\pm$2.9} & 3.96{\small$\pm$0.69} & 1.76{\small$\pm$0.42} & 0.90{\small$\pm$0.03} & 0.15{\small$\pm$0.03} & 0.08{\small$\pm$0.01} \\
    + CM.Contra. & 71.8{\small$\pm$2.3} & 71.1{\small$\pm$2.3} & 70.9{\small$\pm$1.9} & 3.43{\small$\pm$0.01} & 2.23{\small$\pm$0.02} & 0.93{\small$\pm$0.00} & 0.19{\small$\pm$0.00} & 0.08{\small$\pm$0.00} \\
    Ours & 72.7{\small$\pm$1.8} & 73.3{\small$\pm$2.4} & 72.6{\small$\pm$1.7} & 3.05{\small$\pm$0.04} & 2.68{\small$\pm$0.05} & 0.95{\small$\pm$0.00} & 0.22{\small$\pm$0.00} & 0.11{\small$\pm$0.00} \\
    \bottomrule
  \end{tabular}
  }
  \label{tab:abl}
\end{table}

\section{Conclusion}
In this paper, we propose a self-supervised representation learning framework for ultrasound video-speech multi-modal data. To the best of our knowledge, this is the first attempt towards cross-modal representation learning without human annotations for ultrasound data. We designed a simple, yet effective, approach by modelling the affinity between these two modalities. To address the inherent sparse correlation and noise issues in the speech data, we propose a cross-modal contrastive learning scheme and an affinity-aware self-paced learning scheme. Experimental evaluation on two downstream tasks shows that the learned representations can transfer to fetal ultrasound standard plane detection and eye-gaze saliency prediction and improve the average performance accordingly.
The proposed approach shows the potential to mitigate the need for laborious manual annotation work in deep learning-based applications for medical imaging via automated cross-modal self-supervision.
{Since the proposed approach is trained and evaluated on video data with narrative speech audio and is not specifically tailored for ultrasound, it could apply to other unseen video-audio data.}

{
\subsection*{Acknowledgements}
We acknowledge the EPSRC (EP/M013774/1, Project Seebibyte), ERC(ERC-ADG-2015 694581, Project PULSE), and the support of NVIDIA Corporation with the donation of the GPU.
}
%
%
%
\bibliographystyle{splncs04}
\bibliography{mybib}
\end{document}